\documentclass[10pt,twocolumn,letterpaper]{article}
\usepackage[margin=0.72in,columnsep=0.26in]{geometry}
\usepackage[T1]{fontenc}
\usepackage{newtxtext,newtxmath}
\usepackage{microtype}
\usepackage{graphicx,booktabs,tabularx,array,multirow,longtable}
\usepackage[dvipsnames]{xcolor}
\usepackage{amsmath}
\usepackage[numbers,sort&compress]{natbib}
\usepackage{enumitem}
\usepackage{titlesec}
\usepackage{caption}
\usepackage{placeins}
\usepackage{xurl}
\usepackage[colorlinks=true,allcolors=MidnightBlue,breaklinks=true]{hyperref}
\definecolor{ink}{HTML}{162A3A}
\definecolor{accent}{HTML}{087F8C}
\setlist{nosep,leftmargin=*}
\titleformat{\section}{\large\bfseries}{\thesection}{0.6em}{}
\titleformat{\subsection}{\normalsize\bfseries}{\thesubsection}{0.5em}{}
\titlespacing*{\section}{0pt}{9pt plus 2pt minus 1pt}{4pt}
\titlespacing*{\subsection}{0pt}{7pt plus 1pt}{3pt}

\newcommand{\prob}{p_{\theta}}
\newcommand{\figfile}[2]{\includegraphics[width=#2]{figures/#1}}
\newcommand{\resource}[2]{\href{#1}{#2}}
\newcolumntype{L}[1]{>{\raggedright\arraybackslash}p{#1}}
\newcolumntype{R}[1]{>{\raggedleft\arraybackslash}p{#1}}
\title{\vspace{-1.1em}\textbf{RoboJev: An Open Vision-Language Judge\\for Robot Success Detection}\vspace{-0.3em}}
\author{Josef Chen\\[2pt]\small Independent Researcher}
\date{October 2026}
\hypersetup{pdftitle={RoboJev: An Open Vision-Language Judge for Robot Success Detection},pdfauthor={Josef Chen},pdfsubject={Open-weight robot success detection and episode-level evaluation},pdfkeywords={robot success detection, vision-language models, robot evaluation, reward models, RoboJev}}
\begin{document}
\twocolumn[\maketitle
\begin{center}
\begin{minipage}{0.94\textwidth}
\textbf{Abstract.}
Many robot manipulation videos do not indicate whether the task was successfully completed. RoboJev-27B is an open-weight vision-language model that takes a task instruction, 10 frames of a manipulation attempt, and (if available) final frames from auxiliary cameras and predicts whether the robot accomplished the task. RoboJev-27B is a variant of JEV-27B-VL, trained with a rank-32 LoRA adapter on the language model only (keeping the vision encoder frozen) using 52,598 episode records. It also uses a source-specific KL penalty that applies only to research data, anchoring predictions closer to the original model on those sources. It uses a calibrated two-token readout that outputs success/failure probabilities in a single forward pass. On 268 held-out episodes of the same robot model (SO-101), RoboJev-27B attains 95.2\% balanced accuracy and 0.993 AUROC. It also generalises to episodes collected by unseen labs/robots (AUROC = 0.890 over 459 episodes) as well as to clearly labelled community-collected episodes (accuracy = 93.3\% over 135 episodes), where the base model attains an AUROC = 0.900 on the unseen-lab/robot subset with the same inputs. It takes 0.26 seconds per episode (median, serial) to infer on a single H100, which is about \$0.19 per 1,000 success predictions at \$2.69 per GPU-hour (not including media processing and serving costs). RoboJev-27B is released together with its inference code, along with a transparent evaluation: 20,628 predictions organized into 29 model/protocol entries are made available to facilitate browsing and to enable reproducible benchmarking of success detection. We also compare against strong API-based frontier VLMs, which take in fewer input frames (4 images from the primary camera only); note that this comparison is not strictly like-for-like because RoboJev-27B is given more frames at inference.
\end{minipage}
\end{center}
\vspace{0.4em}
\begin{minipage}{\textwidth}
\centering
\figfile{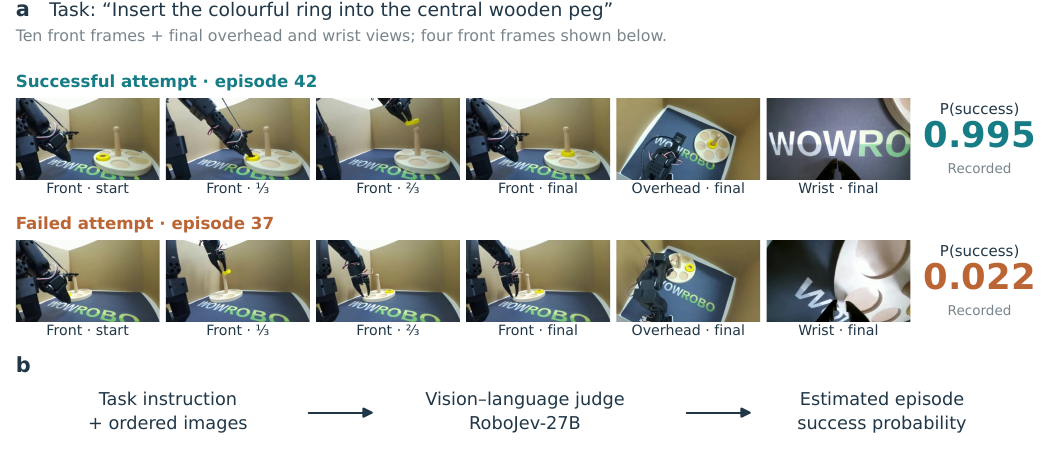}{\textwidth}
\captionof{figure}{\textbf{A task instruction and evidence of an attempt become a success probability.} The real examples of SO-101 ring-placement shown here illustrate the released inference interface. Given a temporal sample, along with optional final auxiliary views, the judge reads them and then scores the two answer tokens. All frames are from ArmnetBench and the figure indicates the exact camera protocol for each score. These examples were selected to illustrate the interface and are not meant to give any estimate of our benchmark performance.}
\label{fig:overview}
\end{minipage}
\vspace{1.0em}
]

\section{Introduction}
\label{sec:intro}
An important ingredient for review queues, learned terminal rewards, or aggregate evaluation is knowing which episodes succeeded. Such labels are often not machine readable and cannot be easily derived from low-level heuristics (e.g. trajectory length, data origin, or smoothness). We therefore fine-tune an episode judge from JEV-27B-VL \citep{jev_model,robojev_model} into RoboJev, an open-weight success detector for robot manipulation videos. Given a task instruction, ten evenly spaced frames from the primary camera, and the final frame from any auxiliary cameras, RoboJev outputs a calibrated success probability. Rather than building a different classifier for every task and environment, RoboJev can be used in a plug-and-play fashion. Rather than generating an explanation, RoboJev scores \texttt{true} / \texttt{false} tokens at a designated position. RoboJev is reusable across tasks and environments thanks to its well-defined inputs, consistent prompting protocol, and verifiable error logs.

We test RoboJev on held-out SO-101 rollouts, on lab/robot domains not included in the robot-specific fine-tuning, and on community videos that have clear labels. Community videos with ambiguous labels can still be inspected, but are not used for binary metrics. While the SO-101 split contains rollouts from the same physical environment as some of the training data, we leave out two entire tasks, as well as individual recording sessions. On this split, RoboJev-27B achieves 95.2\% balanced accuracy. On the unseen-domain data, it achieves 0.890 AUROC, just under the 0.900 AUROC of the unadapted JEV-27B-VL run with the same input protocol. Alongside our released predictions, protocols, and a pinned checkpoint, this indicates that RoboJev is a useful and inspectable tool for policy evaluation, rather than indicating that fine-tuning yields generalizable improvements.

\section{Related work}
\label{sec:related}
SuccessVQA formulates robot outcomes as visual questions and measures how well the model generalizes to unseen instructions and environments \citep{successvqa}. RoboCLIP produces trajectory rewards from a text instruction or video demonstration via a frozen representation \citep{roboclip}. Video-Language Critic trains a transferable reward signal from a dataset of observations using contrastive loss and temporal-ranking loss \citep{videolanguagecritic}. RoboJev uses a supervised label of success/failure for each trajectory and reports a detector that has been publicly released but does not evaluate its use in downstream policy learning.

RoboReward trains general reward models using various robot data, counterfactual instructions, and temporal clips \citep{roboreward}. Robometer employs frame-progress loss, trajectory preference supervision, and releases the RBM-1M dataset \citep{robometer}. Both of these objectives and methods are different from a terminal two-token prediction. Here, we evaluate Robometer-4B using its full video version. We do use the converted data of RoboReward during our training process, but we do not evaluate its model as a direct baseline. ViFailback includes failure trajectories that were originally collected for diagnosis and later correction \citep{vifailback}.

We source data from Open X-Embodiment \citep{oxe} and DROID \citep{droid}, which include demonstration data for diverse tasks and embodiments. Datasets from the community are accessed through the shared format and collection of tools offered by LeRobot \citep{lerobot}. While these datasets allow us to access demonstrations from different embodiments and performing different tasks, they have different viewpoints, success/failure labels, and sources. In Appendix \ref{app:data}, we clearly identify the processed datasets we use in our work, separately from the original datasets.

Our model is based on a pre-trained vision-language model, with low-rank adaptation of the language model component, and a frozen pre-trained vision encoder. We do not propose a novel architecture for vision-language models, but instead leverage off-the-shelf components to build our system. In addition to architectural choices, we use a KL penalty towards the initial model for some data sources, and temperature scaling. Our contributions are an open-weight judge, the full training and inference recipe, and an evaluation artifact describing the trade-offs between specialization, use of visual evidence, and deployment cost.

\begin{figure*}[t]
\centering
\figfile{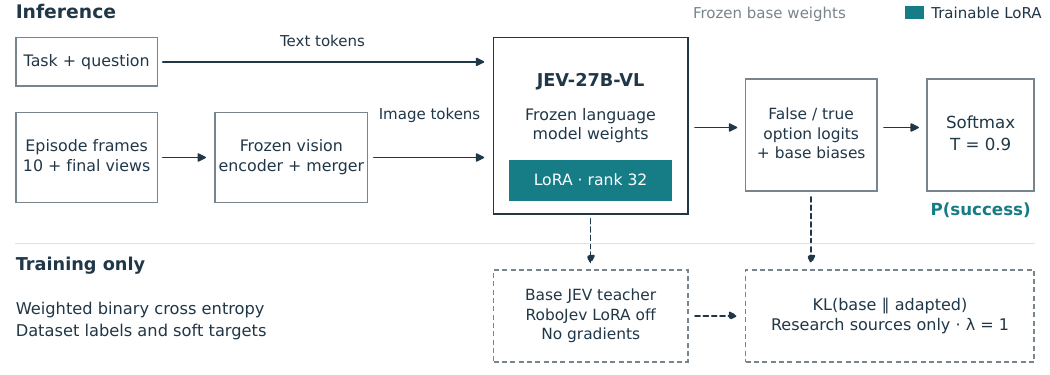}{\textwidth}
\caption{\textbf{An existing vision-language backbone with a task-specific decision adapter.} LoRA (rank 32) is applied only to the language-model component, while the vision encoder and merger remain frozen. Success targets can be binary or soft, and KL anchoring to the base model is applied on a selected subset of research sources. At inference, the merged model scores the two answer tokens with the released calibration applied; the teacher branch is used only during training and is not part of the serving graph.}
\label{fig:architecture}
\end{figure*}

\section{A task-conditioned episode judge}
\label{sec:method}
\subsection{Input and decision readout}
Let $u$ be a task instruction, and $V=\{v_0,\ldots,v_{N-1}\}$ a video captured by the primary camera. Our release samples $K=10$ evenly spaced frames from the beginning to the end of the attempt, and includes the last image from each available auxiliary camera. The prompt specifies the task instruction, chronological order, and all other available views. Success requires completing the instruction by the end of the attempt; a dropped object, incorrect placement, unfinished attempt, or wrong task is negative.

The vision-language model then consumes all this information in one pass, generating the logits $z_0,z_1$ at the answer location for the tokens \texttt{false} and \texttt{true}. We employ the following calibrated readout  \begin{equation}
\prob(y=1\mid u,V)=
\frac{\exp((z_1+b_1)/T)}{\sum_{c\in\{0,1\}}\exp((z_c+b_c)/T)},
\label{eq:readout}
\end{equation} where the release preserves the token IDs, additive biases $b_c$, and the temperature $T$ in \texttt{decision.json}. In the release, we set $T=0.9$. This output provides a measure of the likelihood that the task was successfully completed at the end of the episode; it is not a continuous quantity reflecting the progress made by the agent toward completion of the task, nor does it describe the result in natural language.

\subsection{Training data}
Our model is fit on 52,598 episodes. The data used to fit our model is a mixture of policy failures (from real robots), demonstrations, and simulation interaction data. Specifically, there are 28,170 episodes from RBM-1M sources, 1,388 episodes from ArmnetBench, 10,216 teacher labelled community episodes, 3,718 community demonstration episodes, 4,962 episodes from ViFailback, and 4,144 successful episodes from DROID. These are the number of records used for fitting after the train/valid/test split, and do not necessarily correspond to the number of records in each dataset before the split. We do not claim that these counts are independently de-duplicated. We hold out entire validation datasets, and hold out community data from users that we evaluate on from our training mix. We leave a description of the exact composition of our training set in Appendix \ref{app:data}. There is a small number of records that do not match up with the number returned by the community loader, the cause of which we could not determine.

In total, the community loader reports 11,415 records were accepted, although a subset of these are discarded during validation. We start with annotations that have high-confidence agreement between GPT-6 Astra and GPT-6.1 Sol (probability $\geq 0.9$ or $\leq 0.1$) on the four primary-camera frames. If a Claude Opus 5.5 annotation exists, we discard all annotations it disagrees with on the binary outcome. We then average over all teacher-model annotations to obtain a soft target and discard any examples where the mean is not $\geq 0.9$ or $\leq 0.1$. This process yields a consensus annotation file with 11,387 unique targets. After removing validation examples and those involving held-out robots, 10,216 examples remain for training. To avoid information leakage between training and test labels, we also discard training examples from all 106 owners that appear in the human-labeled community check. The student model still depends on teacher models, which we also use as baselines.

We employ inverse-frequency class balancing for the success/failure examples for each source, clamping weights to $[0.25,4]$, and if a source contains only one type of example, we assign weight 1. For data augmentation we use truncated positive episodes and episodes with mismatched instructions as heuristic negatives. This may introduce label noise because a truncated episode might fulfill a shorter instruction. We also resize, recompress using JPEG, and perturb the colour of images. Eligible episodes with ten-frame evidence use the ten-frame protocol with probability 0.6. The other branches of augmentation preserve the four-frame and final-view examples. In total, we obtain 102,172 training examples in the epoch through data augmentation and repetition.

\subsection{Adaptation and calibration}
We start with JEV-27B-VL, merging in the decision adapter weights before doing any robot-specific adaptation. To adapt LM, we freeze the vision encoder and visual merger, and train LM with rank-32 LoRA \citep{lora} using a learning rate of $5\times10^{-5}$ for one epoch. Denote $q_\theta$ and $q_0$ the two-way softmax outputs of the adapted and original models (before fitted temperature scaling), respectively. Our per-example objective is  \begin{equation}
\mathcal{L}_i = w_i\,\mathrm{CE}(y_i,q_\theta(x_i))
+ \lambda a(x_i)\,D_{\mathrm{KL}}(q_0(x_i)\Vert q_\theta(x_i)).
\label{eq:loss}
\end{equation} The source-specific balancing factor only affects the cross-entropy term, the target distribution of the KL term is still the output distribution of the starting model. For $w_i$, we assign an example weight for each training example: (i) for synthetic negative examples we use a fixed value $w_i=0.5$; (ii) for the original examples, we use source-balanced weights. We apply the anchoring method ($a(x_i)=1$, $\lambda=1$) for training with the RBM/research sources (ViFailback and DROID), but not for community-sourced records or ArmnetBench.

We perform both adapter scaling and calibration on held-out validation sources. Motivated by weight-interpolation \citep{wiseft}, we evaluate the set of scales $\{0,0.25,0.5,0.75,1\}$ on validation data, and select the nonzero scale that gives the highest held-out community AUROC score, while also staying no more than 0.005 AUROC below the base model on each specified validation source. These sources are AutoEval, FINO-Net, the RoboArena 0825 validation split, and the community Koch robots. All of the nonzero DA scales pass this constraint, and scale 1 has the highest community validation AUROC. In this work, AutoEval and FINO-Net are used only for validation; a different split of RoboArena provides training data. For temperature scaling \citep{temperature}, we minimize the negative log-likelihood on validation data for the final-view, four-frame, and ten-frame protocols. Note that multiple protocols can be run on the same episode. It is also worth noting that our published test sets were inspected while creating the released models, so they are not a fully one-shot blind test.

\begin{figure*}[t]
\centering
\figfile{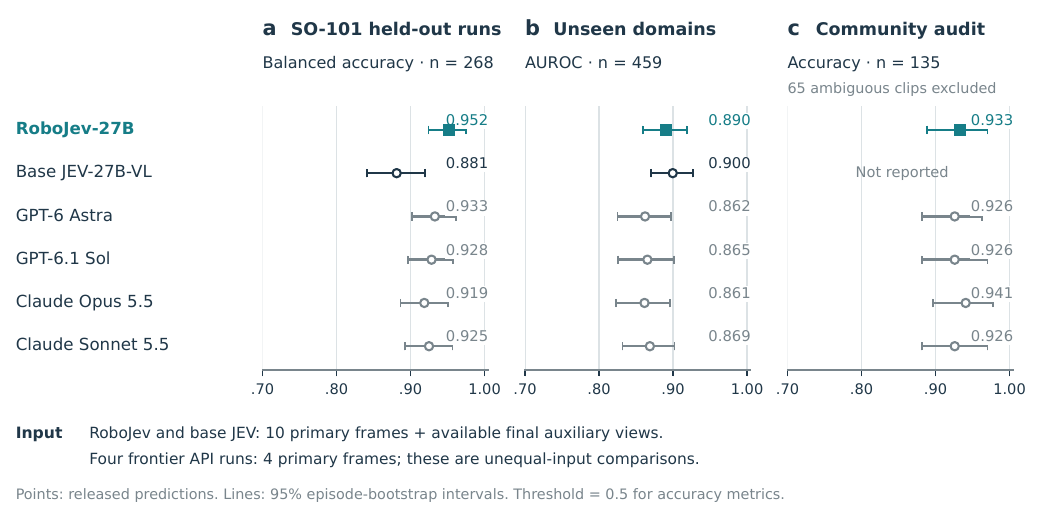}{\textwidth}
\caption{\textbf{Performance depends on the evaluation domain and input protocol.} Shown are results on the held-out SO-101 runs, on the unseen lab/robot domains, and on the clearly labeled community subset. Error bars represent episode bootstrap intervals. All baselines using the API are given the 4 primary camera frames; RoboJev and the dense base receive 10 frames plus available final auxiliary views. There are 458 recorded answers in the unseen-domain plot for Sol. The same-input base achieves a higher unseen-domain AUROC, and Opus achieves higher community accuracy.}
\label{fig:benchmarks}
\end{figure*}

\section{Evaluation protocol}
\label{sec:eval}
\subsection{Three complementary test settings}
There are 927 labelled episodes in the evaluation artifact \citep{robojev_evals}, of which 862 yield a binary success or failure and thus contribute to our reported metrics. The remaining 65 episodes correspond to community videos for which the success or failure is ambiguous. These sets are summarized in Table \ref{tab:splits}.

\begin{table}[t]
\centering\small
\caption{Evaluation composition. ``Held out'' has a different meaning in each setting; the SO-101 task-held-out subset contains 80 of the 268 runs.}
\label{tab:splits}
\begin{tabular}{@{}lrrr@{}}
\toprule
Setting & Success & Failure & Scored\\
\midrule
SO-101 &111&157&268\\
Unseen domains &263&196&459\\
Community &60&75&135\\
\midrule
Total &434&428&862\\
\bottomrule
\end{tabular}
\end{table}

\paragraph{SO-101.} ArmnetBench \citep{armnetbench} contains policy rollouts of 8 tasks and 7 policies. Training and testing recordings are split by session. The tasks \texttt{cable\_unclip} and \texttt{ring\_insert} do not appear in the robot-specific training set at all, and they account for 40 test episodes each. We also report results on the reduced set of held-out tasks, as well as on all 268 test runs, but our overall score is not on entirely unseen tasks.

\paragraph{Unseen lab/robot domains.} The RBM-1M OOD set \citep{rbm_ood} consists of rollouts from domains not present in the robot-specific training set. There are 8 different OOD domains with a total of 459 rollouts. These domains are identified by a combination of the institution, robot (Franka, Koch, Trossen, xArm, SO-101), camera perspective and/or scene condition. The domain sizes range from 16 to 218 rollouts.

\paragraph{Community videos.} The episodes here are from public LeRobot evaluation datasets. One individual labeled each of the 200 episodes, one per dataset. We provide metrics for the 135 episodes with a definite success or failure, and leave the other 65 ambiguous in the artifact. The evaluation is on outcomes visible to human viewers in community videos; we did not evaluate inter-annotator agreement.

\subsection{Model and protocol comparisons}
RoboJev-27B and the ten-frame JEV-27B-VL baseline are evaluated with ten primary-camera frames, as well as final frames from other cameras when available (\texttt{dense-v1} protocol). Recorded frontier API runs use four primary-camera frames corresponding to the beginning, one third, two thirds, and end of the rollout (\texttt{clip-v1} protocol). A Robometer result uses its native video-based interface with the full video. Older RoboJev checkpoints and different adapter scales appear as separate entries.

The released artifact contains 29 model/protocol entries (not 29 distinct architectures), for a total of 20,628 recorded predictions. Some models do not have predictions for all evaluation splits; we treat these as null and exclude them only when computing metrics for the corresponding model, and consider the intersection of answered episodes when comparing models. These results provide useful comparisons with four-frame API models, but do not disentangle the impact of models' capabilities when given the same visual information, since each system follows different protocols.

\subsection{Metrics and uncertainty}
We report AUROC and balanced accuracy, the latter being the mean of sensitivity and specificity. We also report accuracy for the community videos split. For binary predictions, a fixed threshold of 0.5 is used. We report the expected calibration error (ECE) with ten bins. We recompute AUROC and accuracy metrics from all prediction records and are able to reproduce all 118 numbers in the published tables to the reported precision.

We show the matched-input base comparison alongside the frontier API results. These intervals reflect uncertainty due to variability across episodes in each evaluation set, but do not reflect uncertainty due to generalization to new labs, take into account within-domain correlation, or adjust for multiple comparisons. For new results, we use stratified bootstrap with 4,000 replicates (stratified by label) and use percentile 95\% confidence intervals. For comparing paired models, we use the same indices in each bootstrap replicate for both models. Random seed is 20261024. We keep the previously published API intervals for comparability. Those intervals were computed using 4,000 replicates of the (ordinary) bootstrap over episodes. Random seed is 0.

\section{Results}
\label{sec:results}

\begin{figure*}[t]
\centering
\figfile{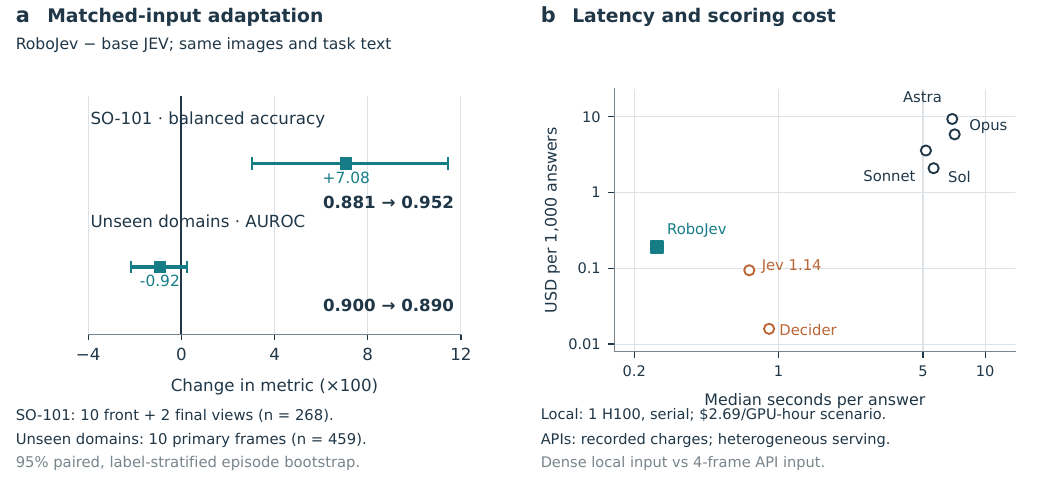}{\textwidth}
\caption{\textbf{Matched-input adaptation and deployment cost.} Left: adapted minus base, paired at the verified identical inputs. The interval includes zero for the unseen domain. Right: Cost per 1,000 answers and response time per answer (medians). API costs are recorded charges; the H100 point uses a GPU-hour estimate. These are not controlled speedups, because the infrastructure and input budget differ.}
\label{fig:adaptation}
\end{figure*}

\subsection{Strong SO-101 discrimination}
Interestingly, the model's performance on held out tasks lags behind some baselines. On the 80 task-held-out episodes from SO-101, RoboJev attains a balanced accuracy of 92.5\% (AUROC 0.986), outperforming the base model using the same inputs (91.3\%, AUROC 0.978). However, four-frame baselines do better here, with four-frame Astra reaching 96.3\% (AUROC 0.987), and four-frame Sol reaching 96.3\% (AUROC 0.998). On the other hand, RoboJev-27B substantially outperforms the base model using the same ten input frames on the 268 held-out runs from the SO-101 dataset, with AUROC 0.993, balanced accuracy 95.2\% (ECE 0.032), and AUROC 0.979, balanced accuracy 88.1\% (ECE 0.114) for the base model (paired difference in balanced accuracy 7.08 percentage points, 95\% interval $[3.04,11.45]$).

GPT-6 Astra (which also provides community training labels) attains 93.3\% balanced accuracy using four frames on the entire held-out SO-101 data. Comparing to this run, RoboJev has a paired balanced-accuracy difference of +1.91 percentage points (95\% interval $[-1.56,5.21]$), which is not resolved as superior as the interval includes zero. We note that there are differences in the visual inputs between our method and the baseline.

\subsection{Transfer reveals the limits of specialization}
When run on the 459 unseen-domain episodes, RoboJev obtains a balanced accuracy of 78.2\% (AUROC 0.890). This is slightly lower than the 79.4\% (AUROC 0.900) obtained by the base model run on the same input episodes (exact same image hashes and task text on all 727 SO-101 and unseen-domain episodes). In our controlled setting, adaptation does not seem to offer any transfer benefit: we find an AUROC difference of $-0.0092$ (95\% difference interval $[-0.0218,0.0026]$).

In a different setting, we compare RoboJev with the recorded four-frame Astra run. This comparison is not controlled for input images, and reflects only a comparison of the recorded protocols. In this comparison, RoboJev obtains an AUROC that is 0.0282 higher (95\% difference interval $[0.0049,0.0514]$). RoboJev's predicted probabilities are not well calibrated after transfer, and thus should not be taken at face value. This is evidenced by RoboJev's ECE on unseen domains, 0.148, which is much higher than on SO-101.

\begin{figure*}[t]
\centering
\figfile{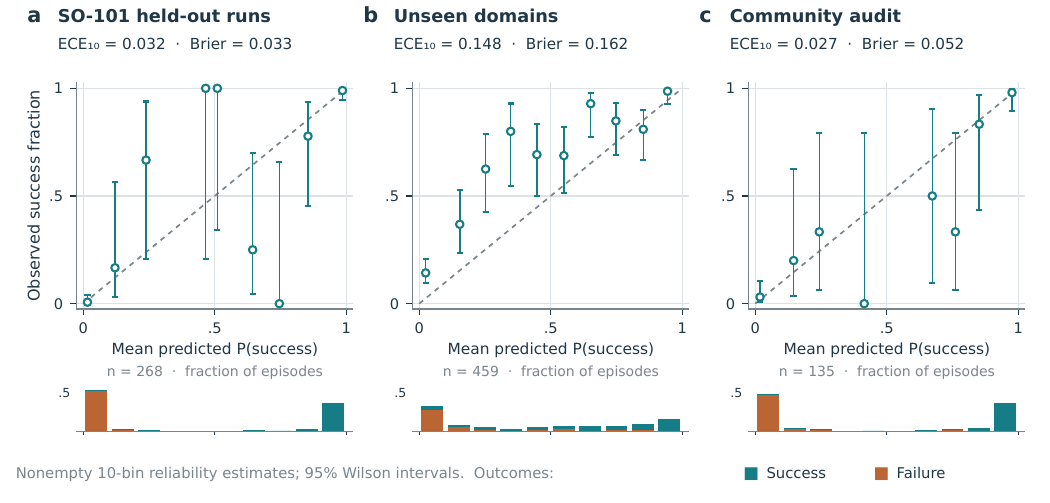}{\textwidth}
\caption{\textbf{Good calibration in one setting does not transfer automatically.} The reliability diagram for the released model uses ten equal-width bins; the vertical error bars represent 95\% Wilson intervals on the observed fraction of successes. In the histograms, we show the fraction of episodes falling into each bin, separated by outcome. The ECE and Brier score are computed over all scored episodes in a given panel. A diagonal line indicates perfect calibration. This represents a post-hoc diagnostic of the evaluation set, rather than another calibration fit.}
\label{fig:calibration}
\end{figure*}

\subsection{Community accuracy and ambiguous outcomes}
RoboJev correctly labels 126 of the 135 (93.3\%) unambiguously-labeled community episodes (AUROC = 0.984), Opus achieves an accuracy of 94.1\%, and Astra and Sol 92.6\%. Thus, the differences in accuracy between methods are small (only one or two episodes). For example, RoboJev and Astra differ in accuracy by 0.74 percentage points (95\% interval $[-4.44,5.93]$).

We do not attempt to extrapolate the accuracies on the unambiguous subset to performance on all community episodes because the 65 ambiguous clips constitute a substantial portion (32.5\%) of the community episodes. As such, we have kept the ambiguous episodes in our released data artifact. We believe this is important as it highlights a case where a detector should not confidently label an unobservable event as ground-truth.

\subsection{Temporal evidence and anchoring}
Relative to the previous four-frame RoboJev release (AUROC 0.987 on SO-101, AUROC 0.872 on the unseen domain), this release obtains AUROC 0.993 and AUROC 0.890, respectively. Note that this is not a controlled comparison (training data and visual input also change). Relative to the four-frame protocol (AUROC 0.880 on the unseen domain), the ten-frame base obtains AUROC 0.900 on the unseen domain. Note that this is not a controlled comparison of the effect of number of frames (prompt wording also changes).

The released model (DA) has higher unseen-domain AUROC than the unanchored variant (DB) (0.890 vs. 0.847), which is consistent with the anchor helping preserve transfer. Note that we report results from a single run for each of these models (i.e., results are not averaged over repeated seeds). Further, we observe that using only the final frame gives noticeably lower performance on unseen domains (0.776 AUROC for the released model). For tasks where the label cannot be inferred from a single image, retaining some temporal context is useful.

\subsection{Calibration changes across domains}
In Figure \ref{fig:calibration}, we show reliability diagrams using ten equal-width probability bins. We observe that calibration worsens across domains. In particular, we see that the ECE increases from 0.032 on SO-101 to 0.148 on the unseen domains, and the corresponding Brier scores are 0.033 and 0.162 respectively. For the clear community subset, ECE is 0.027 and Brier score is 0.052. Some of the bins are quite sparse leading to a wide range for the binomial confidence interval. However, the plot for unseen domain makes it clear that the model often underestimates the true probability by assigning values lower than the success rate of the bin. These plots can motivate the use of local labelled validation when picking a threshold for deploying the classifier to a new domain.

\section{Serving and intended uses}
\label{sec:serving}
The overlay released along with this work computes the score for prefixes of the attempt to show how the score evolves as more evidence is seen. Our LeRobot plugin can be used to compute the episode reward and comes with a processor that outputs a thresholded version of the terminal reward. We also provide a grader that leverages metadata in the LeRobot dataset to figure out the right video offsets and that stores the probability for each episode. Scoring with Eq. \ref{eq:readout} only requires extracting the log probabilities of the two answer tokens. To this end, any standard vision-language inference endpoint can be used to serve the merged checkpoint.

The median response time for serial inference on an H100 is 0.257 seconds, which at a rental rate of \$2.69 per GPU-hour translates to a compute estimate of  \begin{equation}
C_{1000}=1000\,(0.257/3600)\times 2.69
\simeq \$0.192.
\label{eq:cost}
\end{equation} This includes only the GPU compute cost, and not the cost to download and decode the video, queueing delays, time to acquire a GPU, or spare capacity. In addition, the API response times come from an entirely different serving infrastructure and include their own overheads. This is a purely operational measurement rather than a controlled experiment of one piece of hardware vs another. The entire bf16 checkpoint requires around 55 GB of space, and is intended for an 80 GB GPU.

Scaling our measured serial latency, we estimate that scoring each episode in the catalogue would take $\sim$4.9 GPU-hours. The released catalogue samples 68,555 episodes from 7,641 public datasets, totalling 501.17 hours. Note that this is not a report that the released checkpoint has already regraded the catalogue, but an inference-cost scenario computed from the measured latency. As with Eq. \ref{eq:cost}, this estimate excludes media and service costs.

These results support our ability to perform offline evaluation reproducibly as well as review entire episodes post-hoc. We also implemented and tested the terminal-reward integration, but have not yet measured the impact of terminal rewards on policy learning. Using such a signal in an online controller would require additional closed-loop evaluation and a choice of an appropriate operating threshold since a controller can either exploit reward errors or terminate an attempt too early.

\section{Limitations and data responsibility}
\label{sec:limits}
\paragraph{Observability.} All success probabilities depend on what we can see in the given video. If we can't see the object or target (common for wrist-camera videos), or if the video ends too soon to be able to determine success or failure, then we're unable to certify either success or failure. Similarly, due to sparse sampling of frames, it's possible to miss a brief event that might change success or failure, e.g., contact with the target, spillage, or recovery.

\paragraph{Labels and evaluation reuse.} The training labels of the community datasets are model-generated, meaning that agreement filtering might not remove correlated errors among the models. The human labels for our community test videos are produced by one human annotator; we do not have an estimate of the inter-annotator agreement. While we can exclude overlaps with the public sources used in community splits at the level of a named split, we cannot guarantee there are no overlaps with the pretraining data of the underlying foundation model. Finally, related benchmarks were also examined in previous exploratory experiments; thus, we should not consider these intervals as a completely prospective preregistered evaluation.

\paragraph{Domain coverage and calibration.} The reported numbers cover 8 testing domains, and so give limited evidence about how successful the predictor will be in practice. Wrist-camera tasks, and long-horizon tasks like stirring and folding are still difficult to predict. The temperature was calibrated using a validation set. This does not mean that it will be calibrated in other domains. We did not check the detector's sensitivity to instructions being phrased differently, which is a known failure mode of vision language reward models \citep{robormbench}. Neither a low average error nor a high confidence makes the detector a safety monitor.

\paragraph{Licensing and provenance.} The RoboJev checkpoint has been released under the Apache-2.0 license. The training collection combines multiple sources of data, some of which have heterogeneous licenses and provenance requirements. Notably, some of the research datasets that can be accessed via RBM-1M do not have any specific dataset license. Additionally, there are a few community sources that are no longer publicly accessible. We report these missing licenses and sources in lieu of assuming that public availability means we are free to redistribute all underlying assets. For real-robot clips, the paper reports the source, the episode IDs, and the license.

\section{Conclusion}
RoboJev is an open-weight, task-conditioned success detector that, given a task description and episode, performs a single forward pass to return a success probability. On held-out data from SO-101, it attains a balanced accuracy of 95.2\%. Our experiments on unseen domains include a meaningful counterexample where the starting model has a slight edge in ranking outcomes. To enable transparent reuse and encourage further exploration into both failure and success regions, we release the model checkpoint, our predictions on the evaluation dataset, and our inference and evaluation code, making it easy to go through the results on a per-episode basis. Future work will involve extending this comparison to a more controlled setting, by matching inputs to RoboJev. We would also like to include more domains that are labelled by humans, and explore the usage of RoboJev as a learned reward signal for training downstream policies.

\paragraph{Artifacts.} \resource{https://huggingface.co/josefchen/RoboJev-27B}{Weights and inference tools}; \resource{https://huggingface.co/datasets/josefchen/RoboJev-evals}{evaluation records}; \resource{https://huggingface.co/spaces/josefchen/RoboJev-data}{data explorer}; \resource{https://huggingface.co/spaces/josefchen/RoboJev-demo}{interactive demo}.

\paragraph{Writing and verification tools.} Language-model assistants helped with manuscript drafting, source inspection, code used to produce figures, and consistency checks. Candidate edits proposed by StealthGPT were incorporated after being checked against the recorded methods and results. No robot images were generated. Deterministic analysis scripts computed the reported metrics from recorded predictions. The author takes responsibility for the content of the paper.

\FloatBarrier
\bibliographystyle{unsrtnat}
\bibliography{references}

\clearpage
\onecolumn
\appendix
\section{Reproducibility and protocol ledger}
\label{app:repro}
The calibration file and inference code are included with the release. The released model can be found at commit \nolinkurl{e089a9b69a365744b80765057d86583647af95a9} of the repository \resource{https://huggingface.co/josefchen/RoboJev-27B/tree/e089a9b69a365744b80765057d86583647af95a9}{\texttt{josefchen/RoboJev-27B}} (released on 23 October 2026). To further facilitate reproducibility, we provide our code for generating metrics and plots and snapshots of the released benchmark records. The BibTeX entries are also included. We do not release the model weights or private training videos in the artifact.

\begin{table}[h]
\centering\small
\caption{Inference and adaptation settings recorded for the released model.}
\label{tab:settings}
\begin{tabular}{@{}L{0.25\textwidth}L{0.68\textwidth}@{}}
\toprule
Setting & Value and scope\\
\midrule
Starting model & \texttt{autotrust/JEV-27B-VL}; its decision adapter is merged before robot adaptation.\\
Architecture class & \texttt{Qwen3\_5ForConditionalGeneration}; \texttt{model\_type=qwen3\_5} in the pinned checkpoint. The model card's Qwen3.8 naming is not used to infer an independently verified foundation checkpoint lineage.\\
Adapter & Rank 32, language component; frozen vision encoder; full-strength exported merge.\\
Optimization & One epoch, learning rate $5\times10^{-5}$; binary log loss plus source-specific KL anchor of weight 1.0.\\
Input protocol & Ten uniformly spaced primary-camera frames and the last frame of available auxiliary cameras.\\
Answer tokens & \texttt{false}: 3721; \texttt{true}: 1802.\\
Calibration & $T=0.9$; $b_0=+0.0001196011$, $b_1=-0.0001198180$; fixed threshold 0.5 for reported binary metrics.\\
Image preprocessing & 288-pixel short-side target/minimum with JPEG preprocessing; exact client behaviour is part of the release.\\
Serving measurement & One H100, serial requests; median 0.257 seconds per answer; media and allocation overhead excluded.\\
\bottomrule
\end{tabular}
\end{table}

\subsection{Named protocols}
\begin{description}[style=nextline]
\item[\texttt{dense-v1}] Ten evenly spaced frames from the primary-camera video, augmented with the final frames of the other cameras when present; used for the current release and the matched-input base-model row.
\item[\texttt{clip-v1}] Four primary-camera frames at relative positions $0,1/3,2/3,1$; used for the recorded decision-API and frontier baselines and for earlier model variants.
\item[\texttt{done-v1}] Final frame from every available camera; used for the final-frame diagnostic, not the headline release results.
\item[\texttt{video}] The full clip, ingested through a baseline's own video interface; a distinct input budget.
\end{description}

\section{Complete metric inventory}
\label{app:metrics}
Tables are created using the per-episode predictions we have released. There are some entries for which we only released the aggregate performance (see the release summary page) but for the rest we do have the per-episode predictions. The tables are inclusive of all models and protocols (even deprecated ones) for historical comparison. For all tables, each model is evaluated on all episodes for which we have binary labels. Note that a dash does not mean the model had 0 performance, but simply that we don't have predictions recorded for that model/protocol.
\begin{table}[ht]
\centering\small
\caption{Complete available SO-101 and unseen-domain results. BA is balanced accuracy (\%); AUC is AUROC. $n$ is the number of available binary-labelled predictions. D: dense-v1; C: clip-v1; V: whole video.}
\label{tab:all-results}
\setlength{\tabcolsep}{4.5pt}
\begin{tabular}{@{}lcrrrrrr@{}}
\toprule
 & & \multicolumn{3}{c}{SO-101} & \multicolumn{3}{c}{Unseen domains} \\
Model / historical variant & Input & $n$ & AUC & BA & $n$ & AUC & BA \\
\midrule
RoboJev-27B & D & 268 & 0.993 & 95.2 & 459 & 0.890 & 78.2 \\

GPT-6 Astra & C & 268 & 0.987 & 93.3 & 459 & 0.862 & 77.8 \\

GPT-6.1 Sol & C & 268 & 0.989 & 92.8 & 458 & 0.865 & 79.1 \\

Claude Opus 5.5 & C & 268 & 0.978 & 91.9 & 459 & 0.861 & 73.9 \\

Claude Sonnet 5.5 & C & 268 & 0.980 & 92.5 & 459 & 0.869 & 74.4 \\

JEV-27B-VL & C & 268 & 0.983 & 87.7 & 459 & 0.880 & 77.4 \\

TypeSafe Jev 1.14 & C & 268 & 0.975 & 82.9 & 459 & 0.841 & 72.1 \\

Perplexity Decider & C & 268 & 0.954 & 81.1 & 459 & 0.885 & 76.1 \\

Liquid D1 & C & 268 & 0.897 & 82.8 & 459 & 0.828 & 74.7 \\

Mercury Decide & C & 268 & 0.938 & 81.6 & 459 & 0.859 & 75.6 \\

Cloudflare Clef & C & 268 & 0.907 & 77.5 & 459 & 0.803 & 72.0 \\

Cloudflare Clef Flash & C & 268 & 0.810 & 73.6 & 459 & 0.815 & 71.9 \\

Cloudflare Clef Omni & C & 268 & 0.700 & 56.3 & 459 & 0.729 & 58.1 \\

Liquid D1 Ultra & C & 268 & 0.918 & 81.4 & 459 & 0.859 & 73.5 \\

Intern-Decision-4B & C & 268 & 0.772 & 72.9 & 459 & 0.808 & 68.8 \\

RoboJev-27B v1, scale 0.5 & C & 268 & 0.987 & 92.4 & 459 & 0.872 & 76.1 \\

RoboJev-27B historical aggregate & C & -- & -- & -- & -- & -- & -- \\

RoboJev-4B v2 & C & 268 & 0.804 & 66.2 & 459 & 0.635 & 57.3 \\

RoboJev-4B v3, scale 0.25 & C & 268 & 0.950 & 84.0 & 459 & 0.820 & 72.6 \\

RoboJev-4B v3, scale 0.5 & C & 268 & 0.978 & 86.8 & 459 & 0.792 & 69.0 \\

RoboJev-4B v3, scale 0.75 & C & 268 & 0.979 & 89.3 & 459 & 0.773 & 64.1 \\

RoboJev-4B v3, scale 1.0 & C & 268 & 0.978 & 89.1 & 459 & 0.750 & 61.4 \\

GPT-6 Luna Decisions & C & 268 & 0.841 & 74.8 & 459 & 0.736 & 65.1 \\

Perplexity Decider Flash v1.2 & C & 268 & 0.915 & 74.9 & 459 & 0.848 & 73.3 \\

Respan Span-02 & C & 268 & 0.755 & 65.4 & 459 & 0.774 & 67.4 \\

TypeSafe Jev 1.13 & C & 268 & 0.936 & 79.4 & 459 & 0.843 & 72.1 \\

JEV-27B-VL (10 frames) & D & 268 & 0.979 & 88.1 & 459 & 0.900 & 79.4 \\

RoboJev-27B historical dense & D & -- & -- & -- & -- & -- & -- \\

Robometer-4B & V & 268 & 0.777 & 67.7 & 459 & 0.818 & 59.8 \\

\bottomrule
\end{tabular}
\end{table}
\begin{table}[ht]
\centering\small
\caption{Available results on the 135 clearly labelled community episodes. Accuracy is ordinary accuracy (\%), not balanced accuracy. The other 65 community clips are marked unclear.}
\label{tab:community-all}
\begin{tabular}{@{}lcrrr@{}}
\toprule
Model / historical variant & Input & $n$ & AUROC & Accuracy \\
\midrule
RoboJev-27B & D & 135 & 0.984 & 93.3 \\

GPT-6 Astra & C & 135 & 0.991 & 92.6 \\

GPT-6.1 Sol & C & 135 & 0.990 & 92.6 \\

Claude Opus 5.5 & C & 135 & 0.988 & 94.1 \\

Claude Sonnet 5.5 & C & 135 & 0.972 & 92.6 \\

\bottomrule
\end{tabular}
\end{table}

\FloatBarrier
\subsection{The two withheld tasks}
The 80 episodes from \texttt{cable\_unclip} and \texttt{ring\_insert} contain 40 successes and 40 failures. This subset differs from the full 268-run SO-101 evaluation. The frontier rows still use fewer images.
\begin{center}\small
\begin{tabular}{@{}lrr@{}}
\toprule
Model and input & AUROC & Balanced accuracy (\%) \\
\midrule
RoboJev-27B & 0.9862 & 92.50 \\
JEV-27B-VL (10 frames) & 0.9775 & 91.25 \\
GPT-6 Astra & 0.9872 & 96.25 \\
GPT-6.1 Sol & 0.9978 & 96.25 \\
Claude Opus 5.5 & 0.9769 & 92.50 \\
Claude Sonnet 5.5 & 0.9703 & 88.75 \\
\bottomrule
\end{tabular}\end{center}
\subsection{Model identities and primary links}
The links below identify the evaluated service endpoints or upstream checkpoints. They do not verify historical provider routing, billing, or model internals. The recorded Astra request specified medium reasoning effort. The four frontier chat baselines returned a requested numerical probability; RoboJev and decision models use their documented decision readout. Historical RoboJev variants are linked to the release documentation rather than represented as independently downloadable checkpoints.
\begin{longtable}{@{}L{0.38\textwidth}L{0.55\textwidth}@{}}
\caption{Primary model sources. Multiple recorded variants share a source.}\label{tab:model-sources}\\
\toprule Model family / recorded variant & Exact model or endpoint identifier (linked) \\ \midrule\endfirsthead
\toprule Model family / recorded variant & Exact model or endpoint identifier (linked) \\ \midrule\endhead
\bottomrule\endfoot
RoboJev-27B and historical 27B/4B variants & \href{https://huggingface.co/josefchen/RoboJev-27B}{josefchen/RoboJev-27B (release documentation)} \\[2pt]
GPT-6 Astra & \href{https://openrouter.ai/openai/gpt-6-astra}{openai/gpt-6-astra} \\[2pt]
GPT-6.1 Sol & \href{https://openrouter.ai/openai/gpt-6.1-sol}{openai/gpt-6.1-sol} \\[2pt]
Claude Opus 5.5 & \href{https://openrouter.ai/anthropic/claude-opus-5.5}{anthropic/claude-opus-5.5} \\[2pt]
Claude Sonnet 5.5 & \href{https://openrouter.ai/anthropic/claude-sonnet-5.5}{anthropic/claude-sonnet-5.5} \\[2pt]
JEV-27B-VL & \href{https://huggingface.co/autotrust/JEV-27B-VL}{autotrust/JEV-27B-VL} \\[2pt]
TypeSafe Jev 1.14 & \href{https://openrouter.ai/typesafe/jev-1.14}{typesafe/jev-1.14} \\[2pt]
Perplexity Decider & \href{https://openrouter.ai/perplexity/pplx-decider-v1.2}{perplexity/pplx-decider-v1.2} \\[2pt]
Liquid D1 & \href{https://openrouter.ai/liquid/d1}{liquid/d1} \\[2pt]
Mercury Decide & \href{https://openrouter.ai/inception/mercury-decide}{inception/mercury-decide} \\[2pt]
Cloudflare Clef & \href{https://openrouter.ai/cloudflare/clef}{cloudflare/clef} \\[2pt]
cloudflare/clef-flash & \href{https://openrouter.ai/cloudflare/clef-flash}{cloudflare/clef-flash} \\[2pt]
cloudflare/clef-omni & \href{https://openrouter.ai/cloudflare/clef-omni}{cloudflare/clef-omni} \\[2pt]
liquid/d1-ultra & \href{https://openrouter.ai/liquid/d1-ultra}{liquid/d1-ultra} \\[2pt]
open/intern-decision-4b & \href{https://huggingface.co/internlm/Intern-Decision-4B}{internlm/Intern-Decision-4B} \\[2pt]
openai/gpt-6-luna-decisions & \href{https://openrouter.ai/openai/gpt-6-luna-decisions}{openai/gpt-6-luna-decisions} \\[2pt]
perplexity/pplx-decider-flash-v1.2 & \href{https://openrouter.ai/perplexity/pplx-decider-flash-v1.2}{perplexity/pplx-decider-flash-v1.2} \\[2pt]
respan/span-02 & \href{https://openrouter.ai/respan/span-02}{respan/span-02} \\[2pt]
typesafe/jev-1.13 & \href{https://openrouter.ai/typesafe/jev-1.13}{typesafe/jev-1.13} \\[2pt]
open/robometer-4b & \href{https://huggingface.co/robometer/Robometer-4B}{robometer/Robometer-4B} \\[2pt]
\end{longtable}

\FloatBarrier
\section{Data sources and visual provenance}
\label{app:data}
Datasets are labelled as to whether they were used for training judges, for held-out validation, or only for evaluation. Dataset availability and licensing are reported as found upstream. Community episode identifiers and links to the original datasets are available in the public benchmark and dataset catalogue.

\subsection{Training records and validation}
\begin{center}\small
\begin{tabular}{@{}lrr@{}}
\toprule
Source family & Fit records & Local validation / exclusions \\
\midrule
Selected RBM-1M sources & 28,170 & 2,703 validation across all RBM sources \\
ArmnetBench & 1,388 & 139 validation \\
Teacher-labelled community & 10,216 & 917 validation + 282 Koch records \\
Community demonstrations & 3,718 & No local holdout in this loader \\
ViFailback & 4,962 & 224 validation \\
DROID successful conversion & 4,144 & 158 validation \\
\midrule Total fit records & 52,598 & 102,172 examples after repetition/augmentation \\
\bottomrule
\end{tabular}\end{center}
Across all protocols, the native validation outputs contain 4,413 unique episode IDs. There are 11,415 accepted (pre-exclusion) records in the community loader, and we end up with 11,387 unique dataset/episode keys in the target map we save. The 282 held-out Koch records map to 272 unique validation IDs. The cause of the 28-record discrepancy has not been established; it is possible (but not proven) that there are duplicates within a partition with different metadata. We use a deterministic owner/robot split that assigns repeated keys to the same side, so the discrepancy alone does not show train/test contamination.
\paragraph{Training-label provenance.} The loader averages the probabilities returned by the available models, and applies its final confidence margin. Early public release text described teacher labeling as Sol/Opus agreement on ten frames. However, the consensus labeling script that we actually ran and the targets that we saved require high-confidence Sol and Astra labels on four primary-camera frames, and use Opus when available for binary-side agreement and veto. Accordingly, the teachers used for training overlap the API baseline panel. Consistency check: the number of targets saved for the executed run matches the number of values in the final consensus file (11,387), confirming that the executed run read that file.
\paragraph{Optimization details.} Training was carried out with LoRA rank 32, LoRA alpha 64, LoRA dropout 0.05, giving 233,455,616 trainable parameters. We used the AdamW optimizer (weight decay 0) with an initial learning rate of $5\times10^{-5}$, a 5\% warmup followed by cosine decay, gradient-norm clipping at 1.0, per-device batch size 1 and gradient accumulation 2 (global batch 14 on the documented 7-GPU run). The random seed was 7. The source records report a total runtime of around 5.5 hours on 7 H100 GPUs (one of the log files shows only a resumed portion of 41.6 minutes, not the full run). For calibration we searched over temperatures $0.25,0.30,\ldots,5.20$ and selected the value that minimises validation negative log-likelihood.
\subsection{Originating corpora and processed subsets}
The processed RBM-1M repository~\citep{rbm1m,robometer} supplies the exact subset identifiers below. Originating corpora include Open X-Embodiment~\citep{oxe}, MolmoAct~\citep{molmoact}, RoboArena~\citep{roboarena}, RoboReward~\citep{roboreward}, SOAR~\citep{soar}, LIBERO~\citep{libero}, and Meta-World~\citep{metaworld}. AutoEval~\citep{autoeval} and FINO-Net~\citep{finonet} are whole-source validation sets. Failsafe and RACER are attributed through their exact processed-source identifiers; no unverified original-paper attribution is added.
The harvest was not pinned to a complete historical repository revision, which limits byte-for-byte reconstruction of all training inputs. The bundled source ledger records the remote archive paths observed on 24 October 2026, separately from the historic training counts. There are 41 RBM source keys, of which 38 contribute fit records; these are not 41 independent originating datasets.
\begin{longtable}{@{}L{0.73\textwidth}rr@{}}
\caption{Exact processed RBM source identifiers, linked to the verified repository. Fit counts are after local and whole-source holdouts.}\label{tab:rbm-sources}\\
\toprule Processed subset identifier & Fit & Validation \\ \midrule\endfirsthead
\toprule Processed subset identifier & Fit & Validation \\ \midrule\endhead
\bottomrule\endfoot
\small \href{https://huggingface.co/datasets/robometer/processed_datasets}{aliangdw\_metaworld\_metaworld\_eval} & 107 & 11 \\[1pt]
\small \href{https://huggingface.co/datasets/robometer/processed_datasets}{aliangdw\_metaworld\_metaworld\_train} & 97 & 3 \\[1pt]
\small \href{https://huggingface.co/datasets/robometer/processed_datasets}{jesbu1\_auto\_eval\_rfm\_auto\_eval\_rfm} & 0 & 800 \\[1pt]
\small \href{https://huggingface.co/datasets/robometer/processed_datasets}{jesbu1\_failsafe\_rfm\_failsafe} & 600 & 40 \\[1pt]
\small \href{https://huggingface.co/datasets/robometer/processed_datasets}{jesbu1\_fino\_net\_rfm\_fino\_net} & 0 & 229 \\[1pt]
\small \href{https://huggingface.co/datasets/robometer/processed_datasets}{jesbu1\_molmoact\_rfm\_molmoact\_dataset\_household} & 300 & 40 \\[1pt]
\small \href{https://huggingface.co/datasets/robometer/processed_datasets}{jesbu1\_molmoact\_rfm\_molmoact\_dataset\_tabletop} & 300 & 40 \\[1pt]
\small \href{https://huggingface.co/datasets/robometer/processed_datasets}{jesbu1\_oxe\_rfm\_eval\_oxe\_bc\_z\_eval} & 300 & 40 \\[1pt]
\small \href{https://huggingface.co/datasets/robometer/processed_datasets}{jesbu1\_oxe\_rfm\_eval\_oxe\_berkeley\_cable\_routing\_eval} & 163 & 2 \\[1pt]
\small \href{https://huggingface.co/datasets/robometer/processed_datasets}{jesbu1\_oxe\_rfm\_eval\_oxe\_bridge\_v2\_eval} & 300 & 40 \\[1pt]
\small \href{https://huggingface.co/datasets/robometer/processed_datasets}{jesbu1\_oxe\_rfm\_eval\_oxe\_jaco\_play\_eval} & 104 & 5 \\[1pt]
\small \href{https://huggingface.co/datasets/robometer/processed_datasets}{jesbu1\_oxe\_rfm\_eval\_oxe\_toto\_eval} & 98 & 3 \\[1pt]
\small \href{https://huggingface.co/datasets/robometer/processed_datasets}{jesbu1\_oxe\_rfm\_eval\_oxe\_viola\_eval} & 14 & 1 \\[1pt]
\small \href{https://huggingface.co/datasets/robometer/processed_datasets}{jesbu1\_oxe\_rfm\_oxe\_aloha\_mobile} & 256 & 16 \\[1pt]
\small \href{https://huggingface.co/datasets/robometer/processed_datasets}{jesbu1\_oxe\_rfm\_oxe\_austin\_buds\_dataset\_converted\_externally\_to\_rlds} & 45 & 5 \\[1pt]
\small \href{https://huggingface.co/datasets/robometer/processed_datasets}{jesbu1\_oxe\_rfm\_oxe\_berkeley\_cable\_routing} & 300 & 40 \\[1pt]
\small \href{https://huggingface.co/datasets/robometer/processed_datasets}{jesbu1\_oxe\_rfm\_oxe\_berkeley\_fanuc\_manipulation} & 300 & 7 \\[1pt]
\small \href{https://huggingface.co/datasets/robometer/processed_datasets}{jesbu1\_oxe\_rfm\_oxe\_berkeley\_mvp\_converted\_externally\_to\_rlds} & 300 & 20 \\[1pt]
\small \href{https://huggingface.co/datasets/robometer/processed_datasets}{jesbu1\_oxe\_rfm\_oxe\_berkeley\_rpt\_converted\_externally\_to\_rlds} & 300 & 40 \\[1pt]
\small \href{https://huggingface.co/datasets/robometer/processed_datasets}{jesbu1\_oxe\_rfm\_oxe\_dlr\_edan\_shared\_control\_converted\_externally\_to\_rlds} & 45 & 3 \\[1pt]
\small \href{https://huggingface.co/datasets/robometer/processed_datasets}{jesbu1\_oxe\_rfm\_oxe\_furniture\_bench\_dataset\_converted\_externally\_to\_rlds} & 300 & 40 \\[1pt]
\small \href{https://huggingface.co/datasets/robometer/processed_datasets}{jesbu1\_oxe\_rfm\_oxe\_iamlab\_cmu\_pickup\_insert\_converted\_externally\_to\_rlds} & 300 & 21 \\[1pt]
\small \href{https://huggingface.co/datasets/robometer/processed_datasets}{jesbu1\_oxe\_rfm\_oxe\_imperialcollege\_sawyer\_wrist\_cam} & 164 & 4 \\[1pt]
\small \href{https://huggingface.co/datasets/robometer/processed_datasets}{jesbu1\_oxe\_rfm\_oxe\_jaco\_play} & 300 & 32 \\[1pt]
\small \href{https://huggingface.co/datasets/robometer/processed_datasets}{jesbu1\_oxe\_rfm\_oxe\_nyu\_rot\_dataset\_converted\_externally\_to\_rlds} & 14 & 0 \\[1pt]
\small \href{https://huggingface.co/datasets/robometer/processed_datasets}{jesbu1\_oxe\_rfm\_oxe\_stanford\_hydra\_dataset\_converted\_externally\_to\_rlds} & 300 & 30 \\[1pt]
\small \href{https://huggingface.co/datasets/robometer/processed_datasets}{jesbu1\_oxe\_rfm\_oxe\_tokyo\_u\_lsmo\_converted\_externally\_to\_rlds} & 47 & 1 \\[1pt]
\small \href{https://huggingface.co/datasets/robometer/processed_datasets}{jesbu1\_oxe\_rfm\_oxe\_toto} & 300 & 38 \\[1pt]
\small \href{https://huggingface.co/datasets/robometer/processed_datasets}{jesbu1\_oxe\_rfm\_oxe\_ucsd\_kitchen\_dataset\_converted\_externally\_to\_rlds} & 141 & 9 \\[1pt]
\small \href{https://huggingface.co/datasets/robometer/processed_datasets}{jesbu1\_oxe\_rfm\_oxe\_utaustin\_mutex} & 300 & 40 \\[1pt]
\small \href{https://huggingface.co/datasets/robometer/processed_datasets}{jesbu1\_racer\_rfm\_racer\_train} & 600 & 40 \\[1pt]
\small \href{https://huggingface.co/datasets/robometer/processed_datasets}{jesbu1\_roboarena\_0825\_rfm\_roboarena} & 0 & 800 \\[1pt]
\small \href{https://huggingface.co/datasets/robometer/processed_datasets}{jesbu1\_roboarena\_eval\_debug\_nowrist\_roboarena\_eval\_debug\_nowrist} & 4,004 & 40 \\[1pt]
\small \href{https://huggingface.co/datasets/robometer/processed_datasets}{jesbu1\_roboreward\_rfm\_roboreward\_train} & 6,000 & 40 \\[1pt]
\small \href{https://huggingface.co/datasets/robometer/processed_datasets}{jesbu1\_roboreward\_rfm\_roboreward\_val} & 3,971 & 40 \\[1pt]
\small \href{https://huggingface.co/datasets/robometer/processed_datasets}{jesbu1\_soar\_rfm\_soar\_rfm} & 6,000 & 40 \\[1pt]
\small \href{https://huggingface.co/datasets/robometer/processed_datasets}{ykorkmaz\_libero\_failure\_rfm\_libero\_10\_failure} & 300 & 15 \\[1pt]
\small \href{https://huggingface.co/datasets/robometer/processed_datasets}{ykorkmaz\_libero\_failure\_rfm\_libero\_90\_failure} & 300 & 40 \\[1pt]
\small \href{https://huggingface.co/datasets/robometer/processed_datasets}{ykorkmaz\_libero\_failure\_rfm\_libero\_goal\_failure} & 300 & 13 \\[1pt]
\small \href{https://huggingface.co/datasets/robometer/processed_datasets}{ykorkmaz\_libero\_failure\_rfm\_libero\_object\_failure} & 300 & 23 \\[1pt]
\small \href{https://huggingface.co/datasets/robometer/processed_datasets}{ykorkmaz\_libero\_failure\_rfm\_libero\_spatial\_failure} & 300 & 12 \\[1pt]
\end{longtable}
\paragraph{Other source repositories.}
\href{https://huggingface.co/datasets/armnet/armnetbench_v01_robometer}{ArmnetBench}; \href{https://huggingface.co/datasets/josefchen/RoboJev-evals}{Teacher-labelled LeRobot community evaluations}; \href{https://github.com/huggingface/lerobot}{Community demonstrations}; \href{https://huggingface.co/datasets/sii-rhos-ai/vifailback-dataset-lerobot}{ViFailback}; \href{https://huggingface.co/datasets/jnogga/droid_success}{DROID successful episodes}.
The community evaluation and demonstration training source lists are not fully exposed by the released evaluation dataset. The supplied ledger therefore identifies the available provenance without representing it as a complete public training manifest. The public benchmark itself retains source dataset links for its evaluation episodes.
\subsection{Licences and figure credits}
Accessed through RBM-1M, copies of the SOAR, AutoEval, and FINO-Net datasets don't have an explicit licence according to the release audit. ArmnetBench and model weights: Apache-2.0; RoboArena, LIBERO, Meta-World, ViFailback: MIT; RoboReward, MolmoAct, DROID: CC BY 4.0; Open X-Embodiment: source-specific terms. The source bundle includes the paper's figures and documented Armnet frames, not the training corpus. Community source availability has changed, and the complete licence chain for training data isn't fully documented, so provenance remains a limitation.
Figure~\ref{fig:overview} uses \href{https://huggingface.co/datasets/armnet/armnetbench_v01_lerobot_so101}{ArmnetBench SO-101} ring-placement episodes 42 (success) and 37 (failure), credited to Armnet under Apache-2.0. The local source snapshot records revision \nolinkurl{1db69c82b589c7172ee31a6b8f7ef58d1fc2c663}; the historical shard was not re-downloaded for this paper. The extracted input JPEGs are the exact recorded model inputs. The three-camera probabilities are 0.995478816 and 0.022072101 at the pinned RoboJev revision. The public front-only demo receipts for the same clips are different, 0.988252938 and 0.026133848. The figure uses the three-camera scores consistently.
Each figure has a corresponding entry in the bundled source manifest that includes a SHA-256 hash of every frame shown, the public dataset URL and episode index, video offsets, model revision, and query identifier. We do not include generated robot imagery or synthetic experimental screenshots in the figures.

\FloatBarrier
\section{Verification and scope of the release}
\label{app:verification}
The source code for computing metrics in the source bundle computes the reported metrics by re-computing on the per episode records that have been released, entirely on CPU. This confirms the correctness of arithmetic and coverage over those records rather than new inference at scale on 27B or whether every historical API call was accounted for. The interactive demo logs the specific pinned version of weights being used as well as the hashes of any example clips provided. Examples logged by the interactive demo are distinct from any new inference run on a user-uploaded video.

We do not claim state-of-the-art performance over frontier models given the same inputs, well-calibrated probabilities in general, safety certification, a re-evaluated version of the entire community catalogue, or better policies learned from the judge, as these would each require further experiments.

\end{document}